\documentclass[sigconf]{acmart}

\usepackage{algorithmic}
\usepackage{graphicx}
\usepackage{multirow}

\usepackage{amsmath,amssymb,amsfonts}
\usepackage{subfig}
\usepackage{arydshln}
\usepackage[shortlabels]{enumitem}%
\usepackage{appendix}
\usepackage{chngcntr}
\usepackage{xcolor}
\usepackage[table]{xcolor}
\usepackage{booktabs}
\usepackage{xspace}
\usepackage{balance}
\usepackage{tcolorbox}
\usepackage{array}
\definecolor{blueblack}{HTML}{000080}

\usepackage[linesnumbered,vlined,ruled]{algorithm2e}
\newcommand{\method}{\textsc{ReBrain}\xspace}

\AtBeginDocument{%
  }

\setcopyright{none}
\acmYear{2027}
\acmDOI{}
\acmISBN{}
\acmConference[KDD '27]{Proceedings of the 33rd ACM SIGKDD Conference on Knowledge Discovery and Data Mining}{August 1--5, 2027}{San Jose, CA, USA}
\acmBooktitle{Proceedings of the 33rd ACM SIGKDD Conference on Knowledge Discovery and Data Mining (KDD '27), August 1--5, 2027, San Jose, CA, USA}

\begin{document}

\title{iEEG Can Help EEG: Improving Clinical Scalp EEG Analysis through Intracranial Priors}

\author{Xianglong Bao}
\affiliation{%
  \institution{Griffith University}
  \department{School of Information and Communication Technology}
  \city{Brisbane}
  \country{Australia}
}
\email{xianglong.bao@griffithuni.edu.au}

\author{Yi Ding}
\affiliation{%
  \institution{Nanyang Technological University}
  \department{College of Computing and Data Science}
  \country{Singapore}
}
\email{ding.yi@ntu.edu.sg}

\author{Zengxiang Li}
\affiliation{%
  \institution{SingHealth Duke-NUS AI in Medicine Institute}
  \country{Singapore}
}
\email{shawn.li.zx@singhealth.com.sg}

\author{Nan Liu}
\affiliation{%
  \institution{National University of Singapore}
  \department{Duke-NUS Medical School}
  \country{Singapore}
}
\email{liu.nan@duke-nus.edu.sg}

\author{Kewen Wang}
\affiliation{%
  \institution{Griffith University}
  \department{School of Information and Communication Technology}
  \city{Brisbane}
  \country{Australia}
}
\email{k.wang@griffith.edu.au}

\author{Zhe Wang}
\authornote{Corresponding author.}
\affiliation{%
  \institution{Griffith University}
  \department{School of Information and Communication Technology}
  \city{Brisbane}
  \country{Australia}
}
\email{zhe.wang@griffith.edu.au}

\author{Ming Jin}
\authornotemark[1]
\affiliation{%
  \institution{Griffith University}
  \department{School of Information and Communication Technology}
  \city{Brisbane}
  \country{Australia}
}
\email{mingjinedu@gmail.com}

\renewcommand{\shortauthors}{}

\begin{abstract}
Scalp electroencephalography (EEG) is a cornerstone of clinical brain monitoring, making the quality of its signal fundamental to downstream tasks and models. However, that signal is heavily contaminated by non-neural interference and blurred by volume conduction. Existing methods operate almost exclusively within the scalp modality, so they inherit the contamination they would need to overcome. In practice, intracranial EEG (iEEG) observes the neural activity with far higher fidelity, yet it is invasive and therefore scarce, and the two modalities are rarely recorded as pairs at scale, share no common channel space, and offer no clean ground truth to directly supervise a mapping. Motivated by the neuronal synchronization mechanism shared by the two modalities, we propose a reprogramming approach that fits a signal-to-signal model on iEEG alone and freezes it as a cross-modal prior.
We introduce \textbf{\method}, a framework to transfer an iEEG-learned prior to scalp EEG, which re-expresses any scalp signal through that prior and can be inserted upstream of any downstream model with no pairing or fine-tuning.
\method consistently improves host models across recordings from multiple hospitals and across architecture families, and our analysis traces this gain to genuine neural structure carried by iEEG rather than to the architecture itself. This work provides evidence that a scarce clean modality can serve as a transferable prior for an abundant noisy one, offering a general recipe for cross-modal transfer among biosignals. Our code is available at \textcolor{ACMDarkBlue}{\url{https://github.com/Xianglong-Bao/ReBrain}}.
\end{abstract}




\maketitle

\section{Introduction}

Scalp electroencephalography (EEG) is a widely used non-invasive electrophysiological recording of human brain activity, gathered at scale across clinical monitoring and brain-computer interfaces. The signal it records, however, is a degraded observation of cortical activity. Volume conduction through the skull and scalp blurs the underlying neural sources~\cite{nunez2006electric}, and extracranial nuisance from the eyes, muscles, and recording environment overlaps the neural content~\cite{uriguen2015eeg}. Intracranial EEG (iEEG), recorded from electrodes implanted inside the skull, observes this neural activity with much higher fidelity and far less contamination, because the electrode sits close to the active sources and bypasses the dominant scalp pathways~\cite{ball2009signal}. The difficulty is that iEEG is invasive and therefore scarce. It is acquired almost exclusively during the presurgical evaluation of drug-resistant epilepsy~\cite{parvizi2018promises}, and so cannot be collected on a scale comparable to scalp EEG. The two modalities therefore stand in an asymmetry in which one is clean but scarce and the other abundant but noisy, while both are observations of the same kind of neural activity~\cite{buzsaki2012origin}.

This asymmetry raises three scientific questions that organize our study. \textbf{(SQ1)}~Since scalp EEG and iEEG arise from a shared neuronal synchronization mechanism, can iEEG help noisy scalp EEG? \textbf{(SQ2)}~How can this help be realized, and can a simple and effective method use an iEEG-learned prior to process scalp EEG? \textbf{(SQ3)}~What is the mechanism behind this help, that is, what does the prior do to the scalp signal?
Answering them is difficult because several obstacles rule out a direct transfer. A supervised cross-modal mapping has almost no training pairs, because a paired recording would require implanting intracranial electrodes and placing scalp electrodes on the same patient at once, which is rarely done at scale. The two modalities also do not share a channel space, since intracranial montages are patient-specific while scalp montages vary across datasets, so no fixed channel correspondence can be defined. And scalp EEG has no clean ground truth, because any recovered ``clean'' signal is still a transformed observation of the neural process rather than the process itself, so a denoiser has nothing to be supervised against or verified with. Whatever transfers must therefore be carried by iEEG alone, without pairing or ground truth.

Existing approaches leave this gap open.
Self-supervised representation learning on EEG pretrains on the scalp modality itself~\cite{banville2021selfeeg, kostas2021bendrusingtransformerscontrastive, jiang2024biot, jiang2024labram, wang2024eegpt, wang2025cbramodcrisscrossbrainfoundation, el2026reve, zhou2026csbrain}, so it inherits the contamination it would need to overcome and yields task representations rather than a signal that an existing model can consume.
Signal-level denoising instead operates on the scalp recording and targets specific artifact classes, learning from ``clean'' and noisy signal pairs that are synthesized rather than measured~\cite{zhang2021eegdenoisenet, chuang2025art, huang2025eegdfus, choi2025targetedadversarialdenoisingautoencoders}. Its ``clean'' target is only presumed clean, because denoising the scalp signal is a blind source separation (BSS) problem~\cite{jung2000blind, choi2025targetedadversarialdenoisingautoencoders} in which the neural content cannot be verifiably separated from its nuisance, leaving the method to optimize a proxy objective. And even a perfectly separated target would still be the neural process blurred by volume conduction~\cite{nunez2006electric} rather than the process itself. Operating entirely within the scalp modality, it cannot re-express its input through structure learned from a cleaner one. Neither draws on iEEG as the source of what is transferred.

\begin{figure}[t]
    \centering

    \includegraphics[width=\linewidth]{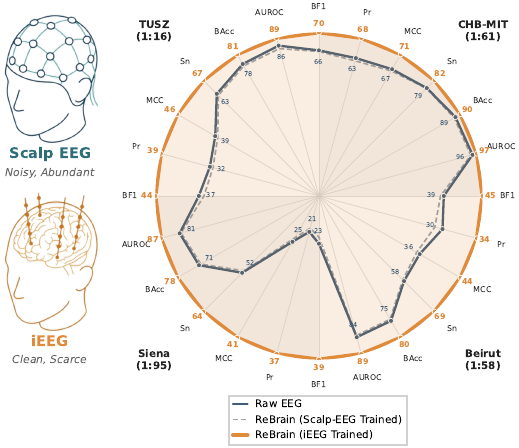}

    \caption{Reprogramming scalp EEG through the frozen iEEG prior improves downstream performance on recordings from multiple hospitals that retain their native clinical class imbalance.}
    \label{fig:teaser}
\end{figure}

We take iEEG itself as that source, and give a deliberately simple answer to how this help can be realized. We fit a signal-to-signal model on iEEG alone and freeze it, then insert it upstream of any downstream scalp-EEG model as a fixed transformation. Within the model, each signal is routed through a finite discrete codebook. Because that codebook is fit on iEEG and has finite capacity, every input processed by the model must be re-expressed as a sequence of cortical-signal atoms drawn from an iEEG-learned vocabulary, and content inconsistent with that vocabulary receives no efficient encoding and is attenuated on reconstruction. Applied to a scalp EEG signal, the frozen model does not filter named artifacts but re-expresses it through the iEEG-defined representation. We call this operation \emph{cross-modal reprogramming} and the frozen model an unpaired cross-modal prior, and we instantiate it as \method, a channel-agnostic model that reprograms each channel independently and so applies unchanged across the heterogeneous montages of both modalities. As Figure~\ref{fig:teaser} shows, this improves the host on recordings from all four hospitals, whereas the same model fitted on scalp EEG instead of iEEG does not, placing the gain in the intracranial corpus rather than the architecture itself.

Our contributions are as follows.
\begin{itemize}

\item We show, through comprehensive experiments, that reprogramming through the iEEG-fitted prior yields consistent, statistically significant gains across recordings from multiple hospitals and downstream models spanning multiple architecture families, under both subject-dependent and subject-independent evaluation.

\item We instantiate the prior as \method, a frozen, channel-agnostic codebook transform that serves as a plug-and-play front-end upstream of any downstream EEG model, requiring no paired data, no labels, and no fine-tuning to deploy.
\item We conduct a detailed mechanistic analysis of what reprogramming does to the scalp signal, and trace the gain to genuine neural structure in iEEG rather than to the architecture itself.
\end{itemize}

\section{Related Work}
\label{sec:related}

\subsection{Signal-Level EEG Enhancement}

Signal-level enhancement operates directly on the scalp recording. Classical pipelines decompose a multi-channel recording into components and discard the ones judged artifactual~\cite{jung2000blind, hyvarinen2000independent, delorme2007enhanced}, or suppress interference by shrinking a wavelet expansion~\cite{donoho1994ideal} and by rejecting segments whose variance exceeds a calibration statistic~\cite{mullen2015real}. Deep denoisers replace these rules with a network trained to regress a clean recording from a corrupted one~\cite{zhang2021eegdenoisenet, chuang2025art, huang2025eegdfus, choi2025targetedadversarialdenoisingautoencoders}.

Both face a supervision problem that the scalp modality cannot resolve. The clean targets a deep denoiser regresses toward are synthesized by adding artifact templates to recordings presumed clean, so the network learns the corruption process its designer wrote down rather than the one the scalp imposes. Classical decompositions need no such templates, but the separation they perform is blind~\cite{jung2000blind} and its solution unobservable, which leaves the labelling of components to a heuristic that no ground truth can check. Underneath both lies the fact that even an exact separation would return cortical activity blurred by volume conduction rather than the activity itself~\cite{nunez2006electric, kayser2015surface}, so no clean scalp target exists to be verified against. Confined to the scalp modality throughout, these methods cannot import structure from a modality that observes the process more directly.

\subsection{Cross-Modal Transfer and Model Reprogramming}

Transfer across biosignal modalities is typically built on paired recordings. Some work uses diffusion models to translate one cardiovascular waveform into another, generating the electrocardiogram from photoplethysmography~\cite{shome2024rddm, chen2025unicardio}, and others align the intermediate representations of foundation models from different physiological modalities through a lightweight bridge~\cite{li2026bioxbridge}. In every such method the transfer is learned from the two modalities recorded together, which is why existing work names the unpaired setting as an open problem~\cite{li2026bioxbridge}. For iEEG and scalp EEG, however, such paired recording is rarely collected at scale, since it would require intracranial implantation and scalp placement within the same session.
Neural codecs originally developed for audio~\cite{defossez2022encodec} have been adapted to brain signals, where iEEG yields better compression than scalp EEG~\cite{carzaniga2025braincodec}. That work targets reconstruction fidelity and overlooks whether such cross-modal information could serve as a prior for downstream tasks.
Our operation is closer in spirit to model reprogramming, which repurposes a network by transforming its input~\cite{elsayed2019adversarial}, and which crosses modalities to fit time series to an acoustic model~\cite{yang2021voice2series} or to a language model~\cite{jin2024timellm}. There the target data is transformed so that a frozen source model can perform a new task, and that source model produces the answer. We reverse these roles. Our frozen source-modality model yields neither an answer nor a representation but a signal in the target domain, which any downstream model then consumes on its own terms. Underwriting this is the view of a fitted network as an implicit prior over signals, established for images by architectures that regularize reconstruction without training data~\cite{ulyanov2018deepimageprior} and by denoisers reused as priors inside unrelated inverse problems~\cite{venkatakrishnan2013ppp}. We carry such a prior across an observation modality rather than across a task, so the structure it imposes is learned from a more direct view of the same underlying process.

\section{Problem Setup and Cross-Modal Asymmetry}
\label{sec:setup}

\begin{figure*}[t]
    \centering
    \includegraphics[width=\linewidth]{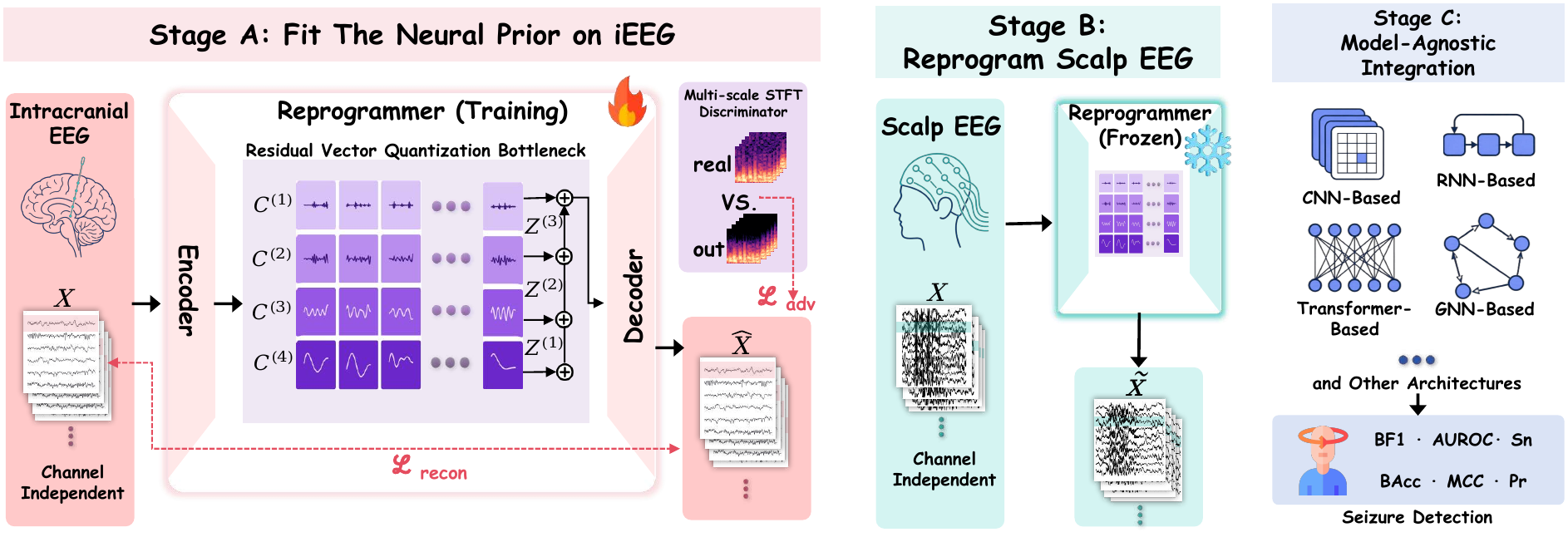}
    \caption{Overview of \method. \textbf{Stage A} fits the neural prior on iEEG alone. \textbf{Stage B} freezes this prior and reprograms scalp EEG through it, re-expressing the raw signal in the iEEG-learned representation. \textbf{Stage C} feeds the reprogrammed signal to a downstream model of any architecture.}
    \label{fig:framework}
\end{figure*}
We consider human brain neural activity as an underlying latent process $s(t) \in \mathbb{R}^{N_s}$ indexed by continuous time $t$, where $N_s$ denotes the effective number of neural sources. This same process can be observed through two distinct electrophysiological modalities that share the latent $s(t)$ but differ markedly in their physical forward operators and contamination structure~\cite{buzsaki2012origin}.

\paragraph{Intracranial observations (iEEG)} Electrodes placed intracranially yield observations
\begin{equation}
x^{\mathrm{ie}}(t) \;=\; A^{\mathrm{ie}}\, s(t),
\label{eq:iEEG-model}
\end{equation}
where $A^{\mathrm{ie}}$ is a near-identity linear operator. The proximity of the electrode to active cortical sources and the absence of intervening tissue produce observations with high signal-to-noise ratio and rich neural structure~\cite{parvizi2018promises}. Moreover, $x^{\mathrm{ie}}(t)$ is substantially less affected by the contaminants of the scalp recording environment, whose dominant pathways are bypassed by the intracranial recording site~\cite{ball2009signal}. However, clinical human iEEG is acquired almost exclusively during the presurgical evaluation of drug-resistant epilepsy, which inherently limits the scale at which such corpora can be assembled~\cite{parvizi2018promises, zijlmans2019presurgical}.

\paragraph{Scalp observations (EEG)} Surface electrodes placed on the scalp instead yield
\begin{equation}
x^{\mathrm{ee}}(t) \;=\; A^{\mathrm{ee}}\, s(t) \;+\; \zeta^{\mathrm{ex}}(t),
\label{eq:EEG-model}
\end{equation}
where $A^{\mathrm{ee}}$ is an operator implementing volume conduction through the intervening tissue layers between cortex and scalp~\cite{nunez2006electric}. The term $\zeta^{\mathrm{ex}}(t)$ denotes a composite extracranial nuisance signal that aggregates the various contaminants introduced by the scalp recording environment~\cite{uriguen2015eeg}, which are substantially attenuated in intracranial recordings~\cite{ball2009signal}. Scalp EEG is non-invasive and far more widely recorded across clinical and non-clinical settings.

\paragraph{The cross-modal asymmetry.} Beyond this accessibility gap, the two modalities differ along several structural axes that frame the central challenge of this work. \emph{Heterogeneity}: channel configurations are heterogeneous both within and across modalities. iEEG montages are patient-specific in electrode count and anatomical placement, while scalp EEG, despite the standardized 10--20 and 10--10 electrode-position conventions, varies considerably across datasets in both the chosen electrode subset and the reference scheme adopted. The two modalities therefore do not share a common channel space onto which a direct correspondence could be defined. \emph{Absence of pairing}: same-patient simultaneous iEEG--EEG recordings require both intracranial electrode placement and concurrent scalp electrode placement, and are rarely carried out at scale, so a supervised cross-modal mapping cannot be trained. \emph{Absence of ground truth}: Standalone EEG denoising approaches operate only on scalp recordings, and any ``clean EEG'' they produce remains a transformed observation of $s(t)$ rather than $s(t)$ itself, so it cannot serve as a faithful ground truth for the underlying neural process. \emph{Ill-posedness}: the scalp forward map $A^{\mathrm{ee}}$ is non-injective, precluding the exact recovery of $s(t)$ from $x^{\mathrm{ee}}$. In addition, the composite nuisance $\zeta^{\mathrm{ex}}$ cannot be explicitly separated from $A^{\mathrm{ee}} s(t)$ in $x^{\mathrm{ee}}$. The task therefore takes the form of a blind source separation (BSS) problem in which the underlying decomposition is itself unobservable. For the same reason, EEG denoising methods cannot directly verify the extent to which they have separated the two.

Taken together, these asymmetries rule out modality-aligned paired training, supervised denoising on clean scalp targets, and approach that requires explicit inversion of the volume-conduction forward map. They define the design space within which a cross-modal approach must operate.

\paragraph{Problem statement.} Our central question is whether iEEG, despite never being seen by downstream models, can serve as a transferable neural prior for models operating on scalp EEG. We instantiate this as the search for a parametric function $f_\theta : \mathbb{R}^{L} \rightarrow \mathbb{R}^{L}$ that re-expresses a raw brain-signal segment $x \in \mathbb{R}^{L}$ through a prior learned from iEEG, mapping it to an output segment $\tilde{x} = f_\theta(x)$ of the same length $L$, with three defining properties. First, $f_\theta$ is fitted exclusively on the iEEG signal itself. No scalp EEG, no downstream label, and no auxiliary supervision enter the training process, making the prior task-agnostic and label-agnostic. Second, once fitted, $f_\theta$ is frozen and applied as a fixed transformation at deployment, requiring no fine-tuning when crossing to scalp EEG or when switching downstream tasks or models. Third, $f_\theta$ operates on signal-level rather than representation-level. It maps a raw signal segment to one of the same shape, so it can be inserted directly upstream of any downstream model $g_\phi$ as a plug-and-play frontend, without architecture change or feature-space bridging.

\section{Method}
\label{sec:method}

As illustrated in Figure~\ref{fig:framework}, we instantiate $f_\theta$ as \method, an encoder-decoder network with a discrete latent bottleneck. An encoder $E$ parameterizes a categorical posterior distribution $q(z|x)$ over discrete latent indices $z$ given an input brain signal $x$. A finite codebook of embedding vectors serves as the latent embedding table, and the indices drawn from $q(z|x)$ select corresponding entries from this table. A decoder $D$ then reconstructs the signal from the selected embeddings, defining a likelihood $p(x|z)$ over input signals. The set of signals expressible through this codebook constitutes the implicit prior of \method on neural waveform structure, which we learn from iEEG and apply to scalp EEG.

After fitting on iEEG, the frozen network re-expresses each scalp EEG signal through this iEEG-defined representation manifold rather than filtering specific artifact types. We call this paradigm \emph{cross-modal reprogramming}.

\subsection{Channel-Agnostic Backbone}
\label{subsec:enc-dec}

$f_\theta$ handles an arbitrary number of channels and time length. We write a multi-channel segment as $X \in \mathbb{R}^{C \times L}$ for $C$ electrodes and $L$ samples, and apply $f_\theta$ independently to each channel $x_c \in \mathbb{R}^L$, stacking the outputs back into a reprogrammed signal of the same shape. The same model thus fits the heterogeneous montages of both modalities without any architectural change.
The encoder $E$ is a 1D convolutional stack of $B$ strided, dilated residual stages that downsample time while expanding features, with total compression $R = \prod_{b=1}^{B} r_b$ producing a latent of length $L/R$ and dimension $d$. The decoder $D$ mirrors it with transposed convolutions.

\subsection{Hierarchical Vector-Quantized Prior Bottleneck}
\label{subsec:rvq}

The latent representation produced by $E$ is passed through a hierarchical residual vector quantizer $Q$, which constrains it to a finite discrete vocabulary via a stack of $M$ codebooks. We adopt $M$ residual codebooks $\{\mathcal{C}^{(m)}\}_{m=1}^{M}$, each containing $K$ entries $\mathcal{C}^{(m)} = \{c_1^{(m)}, \dots, c_K^{(m)}\}$ with $c_k^{(m)} \in \mathbb{R}^d$. Each codebook entry serves as a learned prototype of iEEG signal structure. For each latent vector $z \in \mathbb{R}^d$ in the encoder output, the quantizer iteratively encodes its successive residuals,
\begin{equation}
\begin{aligned}
z^{(1)} &\;=\; z, \\
q^{(m)} &\;=\; \operatorname*{arg\,min}_{c \in \mathcal{C}^{(m)}} \| z^{(m)} - c \|_2, \\
z^{(m+1)} &\;=\; z^{(m)} - q^{(m)}.
\end{aligned}
\label{eq:rvq}
\end{equation}
The quantized latent is the cumulative sum $\hat{z} = \sum_{m=1}^{M} q^{(m)}$, with the discrete tuple $(k_1, \dots, k_M) \in [K]^M$ serving as its index code. Different codebooks thus capture features at different scales of the brain signal, with each layer refining what the previous ones leave behind.
The bottleneck uses $M$ residual codebooks of size $K$, quantizing each latent vector to a budget of $M \log_2 K$ bits. With this factorization, the prior addresses $K^M$ distinct latent configurations using only $M K d$ codebook parameters, achieving exponential combinatorial capacity at linear parameter cost, which would be infeasible to match with a single $K^M$ entry codebook. This finite-bit constraint is the central mechanism through which $f_\theta$ functions as a prior. Every input passing through $f_\theta$ must be re-expressed as a sequence of cortical-signal atoms drawn from the codebook, and components inconsistent with the iEEG-fitted codebook distribution receive no efficient encoding and are implicitly attenuated upon reconstruction.

Codebook entries themselves are initialized by k-means on the first training batch, then trained via exponential moving average (EMA) updates of the encoder activations assigned to each entry, with each $c_k^{(m)}$ set to the running mean of activations routed to it. To prevent the codebook from collapsing onto a small subset of over-used entries, in which case the learned prior loses coverage of the iEEG signal distribution, we reset entries whose EMA usage falls below a threshold. Let $N_k^{(m)}(t)$ denote the EMA cluster size of $c_k^{(m)}$ at training step $t$, updated as
\begin{equation}
N_k^{(m)}(t) \;=\; \alpha\, N_k^{(m)}(t-1) \;+\; (1-\alpha) \sum_{z \in \mathcal{B}_t} \mathbb{1}\!\left[\,q^{(m)}(z) = c_k^{(m)}\,\right],
\label{eq:ema-cluster-size}
\end{equation}
where $\mathcal{B}_t$ is the set of encoder outputs in the current batch and $\alpha \in (0,1)$ is the EMA decay. Entries satisfying $N_k^{(m)}(t) < \tau$ for a threshold $\tau$ are re-initialized by drawing a sample from $\mathcal{B}_t$. This mechanism keeps the codebook well-distributed across the iEEG signal manifold throughout training, preserving the effective coverage of the learned prior.

\subsection{Multi-Scale STFT Discriminator}
\label{subsec:discriminator}

Intracranial recordings contain rich sharp transient structure at fine time-frequency scales, such as interictal spikes and sharp waves. Without an explicit signal valuing this fine-grained content, training can push reconstructions toward lower-frequency averages that dominate aggregate error, losing the transients that define cortical activity. To preserve this structure, we introduce a multi-scale short-time Fourier transform (STFT) discriminator $\mathcal{D}$, adapted from neural codecs~\cite{defossez2022encodec, carzaniga2025braincodec}, that learns to distinguish real iEEG signals from their reconstructions, supplying an adversarial training signal that pushes reconstructions to retain the time-frequency structure of real iEEG.
$\mathcal{D}$ operates on the STFT magnitude representations of the input as a parallel ensemble of $J$ sub-discriminators $\{D_j\}_{j=1}^{J}$, each working at a distinct STFT resolution specified by FFT length $n_{\mathrm{fft}}$, hop length $h$, and window length $w_{\mathrm{win}}$. Each $D_j$ computes a magnitude spectrogram of its input via STFT, then applies a stack of strided 2D convolutions with leaky ReLU activations, producing a 2D map of discrimination logits over the downsampled spectrogram.

This multi-scale design aligns naturally with brain signals, whose informative content spans heterogeneous time-frequency scales. Transient cortical events require fine temporal localization, captured by short-window resolutions, while ongoing oscillatory rhythms from delta to gamma require fine frequency localization, captured by long-window resolutions. As a training-time auxiliary, $\mathcal{D}$ is discarded once training completes and is not part of the frozen prior $f_\theta$ used for downstream EEG inference.

\subsection{Training Objective}
\label{subsec:objective}

We fit $f_\theta$ end-to-end with a composite objective that decomposes into three components. The total generator loss takes the form
\begin{equation}
\mathcal{L}_{\mathrm{gen}} \;=\; \mathcal{L}_{\mathrm{recon}} + \mathcal{L}_{\mathrm{adv}} + \mathcal{L}_{\mathrm{commit}},
\label{eq:gen-loss}
\end{equation}
where $\mathcal{L}_{\mathrm{recon}}$ measures the fidelity of the reconstruction $\hat{x} = f_\theta(x)$ to the input $x$ in time and frequency, $\mathcal{L}_{\mathrm{adv}}$ engages a multi-scale STFT discriminator to push reconstructions toward perceptual realism, and $\mathcal{L}_{\mathrm{commit}}$ anchors encoder outputs to the codebook to stabilize residual-VQ training. The relative weights of the components are not tuned manually. They are dynamically rescaled by a gradient-norm balancer that equalizes per-term contributions to the gradient at the model output. Full expansion of each component is provided in Appendix~\ref{app:loss}.

\subsection{Deploying the Frozen Prior on Scalp EEG}
\label{subsec:deploy}

Once fitted, $f_\theta$ is frozen with all parameters $\theta = (\theta_E, \theta_Q, \theta_D)$ held fixed. At deployment, the frozen network is inserted as a deterministic transformation upstream of any downstream model $g_\phi$,
\begin{equation}
\tilde{x}^{\mathrm{ee}} \;=\; f_\theta\!\left(x^{\mathrm{ee}}\right), \qquad
\hat{y} \;=\; g_\phi\!\left(\tilde{x}^{\mathrm{ee}}\right),
\label{eq:deploy}
\end{equation}
so that $g_\phi$ operates on the reprogrammed signal $\tilde{x}^{\mathrm{ee}}$ rather than on the raw scalp input. The downstream model $g_\phi$ may be any architecture originally designed for raw scalp EEG and is trained using its original protocol and hyperparameters, without any knowledge of $f_\theta$.

\paragraph{Channel-wise forward pass.} Given a multi-channel scalp EEG window $X^{\mathrm{ee}} \in \mathbb{R}^{C \times L}$, we apply Eq.~\eqref{eq:deploy} channel-by-channel in parallel:
\begin{equation}
\tilde{X}^{\mathrm{ee}}_{c, :} \;=\; f_\theta\!\left(X^{\mathrm{ee}}_{c, :}\right), \quad \forall\, c \in [C].
\end{equation}
The reprogrammed tensor $\tilde{X}^{\mathrm{ee}}$ preserves the shape and channel ordering of $X^{\mathrm{ee}}$. Critically, no spatial mixing occurs across channels, so any spatial structure consumed by downstream models is preserved exactly.

\paragraph{Training–deployment decoupling.} The cost of fitting $f_\theta$ is paid once and amortized across all downstream development. Task-level work then proceeds as in standard scalp-EEG pipelines, with $\tilde{X}^{\mathrm{ee}}$ substituting for $X^{\mathrm{ee}}$. This contrasts with paradigms that require joint training, fine-tuning, or task-aware adaptation of the cross-modal component.

\begin{figure*}[t]
    \centering
    \includegraphics[width=\linewidth]{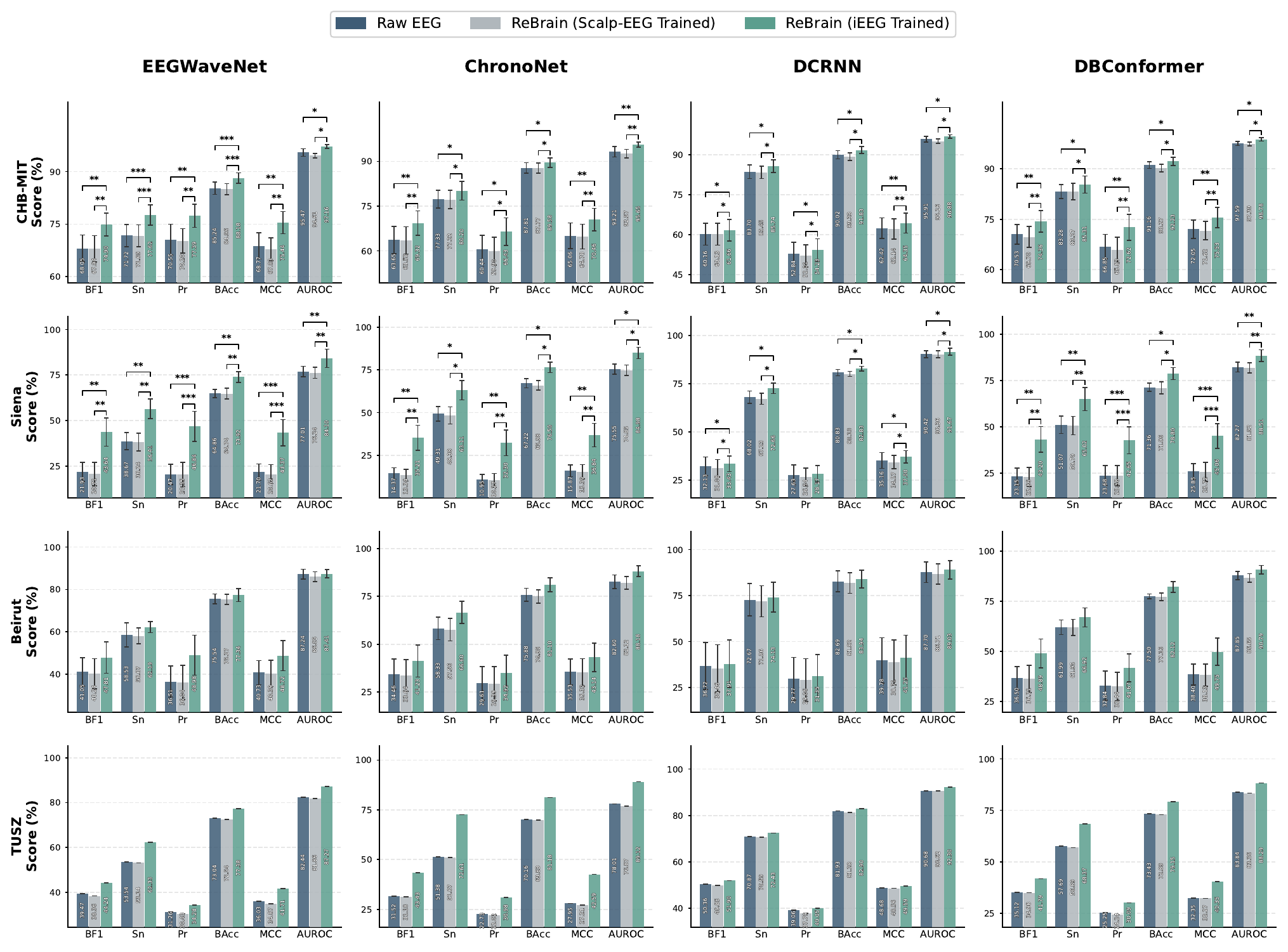}
    \caption{Effect of reprogramming on downstream seizure detection. The three bars per metric are the host model trained on raw scalp EEG, on EEG reprogrammed by a prior fitted to scalp EEG itself, and on EEG reprogrammed by the iEEG-fitted prior. Error bars are the cross-patient standard error and brackets mark per-patient Wilcoxon significance (\(^{*}p<0.05\), \(^{**}p<0.01\), \(^{***}p<0.001\)), both defined only where a per-patient distribution exists. TUSZ uses a subject-independent evaluation split and therefore carries neither error bars nor significance markers. Beirut has only five patients, for which a two-sided signed-rank test cannot reach \(p<0.05\) even if every patient improves, so a consistent effect there carries no marker.}
    \label{fig:main}
\end{figure*}

\section{Experiments}
\label{sec:exp}

\subsection{Experimental Setup}
\label{subsec:setup}

\paragraph{Datasets \& Tasks.} Our study considers the following recordings collected in different hospitals. CHB-MIT~\cite{shoeb2010chbmit} is a pediatric single-center scalp database from Children's Hospital Boston, USA, with 24 recordings from 23 patients. We exclude chb-12, whose channel data are unreadable, together with the repeat recording chb-21, retaining 22 patients~\cite{thuwajit2022eegwavenet}. TUSZ v2.0.3~\cite{shah2018temple} is the largest publicly available clinical seizure EEG corpus to date, from Temple University Hospital, USA, used through its official split of 579 training patients and 43 disjoint evaluation patients. Siena~\cite{detti2020siena} is an adult focal-epilepsy dataset from the University of Siena, Italy, retaining 11 patients after removing three with a single recording. Beirut~\cite{nasreddine2021beirut} is an adult epilepsy dataset from the American University of Beirut Medical Center, Lebanon, retaining 5 patients after removing one missing the Cz and Pz electrodes. SWEC~\cite{burrello2019swec} is an intracranial EEG database from the Sleep-Wake-Epilepsy-Center of Bern University Hospital, Switzerland, comprising 16 patients. All recordings were inspected by an expert neurologist.

Clinical human intracranial EEG is acquired almost exclusively during the presurgical evaluation of drug-resistant epilepsy, where electrodes are surgically placed beneath the skull through craniotomy or stereotactic implantation to localize the seizure focus. We accordingly use seizure detection as our primary task, the clinical setting in which the two modalities most directly overlap, since scalp EEG is a frontline tool for diagnosing and monitoring the very seizures that iEEG is implanted to localize.

\begin{figure}[t]
\centering
\includegraphics[width=\linewidth]{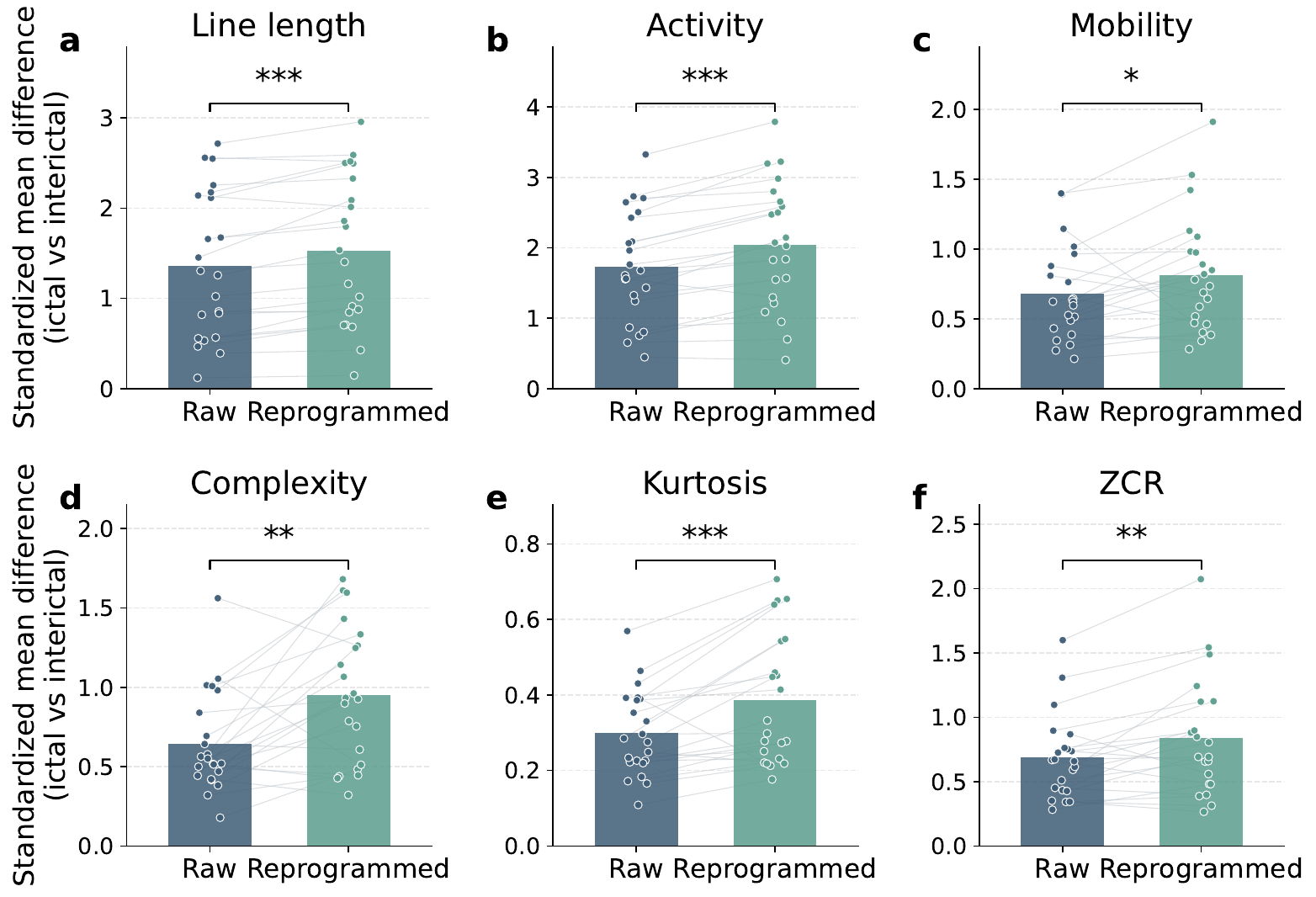}
\caption{Per-feature ictal--interictal separability before and after reprogramming. Each panel is one morphology descriptor and each dot a per-subject standardized mean difference between the two classes, with a subject's raw and reprogrammed values joined.}
\label{fig:perfeat}
\end{figure}

\paragraph{Evaluation protocol.} We consider both subject-dependent and subject-independent evaluation. On CHB-MIT, Siena, and Beirut we run subject-dependent leave-one-recording-out cross-validation, holding out one full recording at a time and training on the remaining recordings of the same patient, so that no window from the held-out recording can leak into training. This yields a per-patient score distribution that supports cross-patient significance testing. TUSZ instead provides a predefined patient-level train and test split, on which we adopt the subject-independent setting, training and testing on disjoint patients to measure cross-subject generalization. It therefore produces a single pooled score per metric rather than a per-patient distribution.

\paragraph{Metrics.} Seizure data is heavily imbalanced, with interictal background windows far outnumbering ictal windows. Under this imbalance conventional metrics such as accuracy approach 100\% even for a weak detector that mostly predicts background. We therefore report six complementary metrics that stay informative under imbalance, namely binary F1 on the ictal class, sensitivity, precision, balanced accuracy, the Matthews correlation coefficient (MCC), and AUROC. None of these metrics is dominated by the majority background class, so their values are substantially lower than that near-perfect accuracy.

\paragraph{Host architectures.} Reprogramming is model-agnostic, so we test it as a drop-in front-end for classic scalp-EEG models that span the major architecture families. EEGWaveNet~\cite{thuwajit2022eegwavenet} is CNN-based, ChronoNet~\cite{roy2019chrononet} is RNN-based, and DBConformer~\cite{dbconformer} is transformer-based, all operating on the time-domain waveform, while the GNN-based DCRNN~\cite{tang2022dcrnn} instead consumes a per-window spectral representation over a learned channel graph. Each host is trained with its own original protocol and is never aware of \method.

\begin{figure}[t]
\centering
\includegraphics[width=\linewidth]{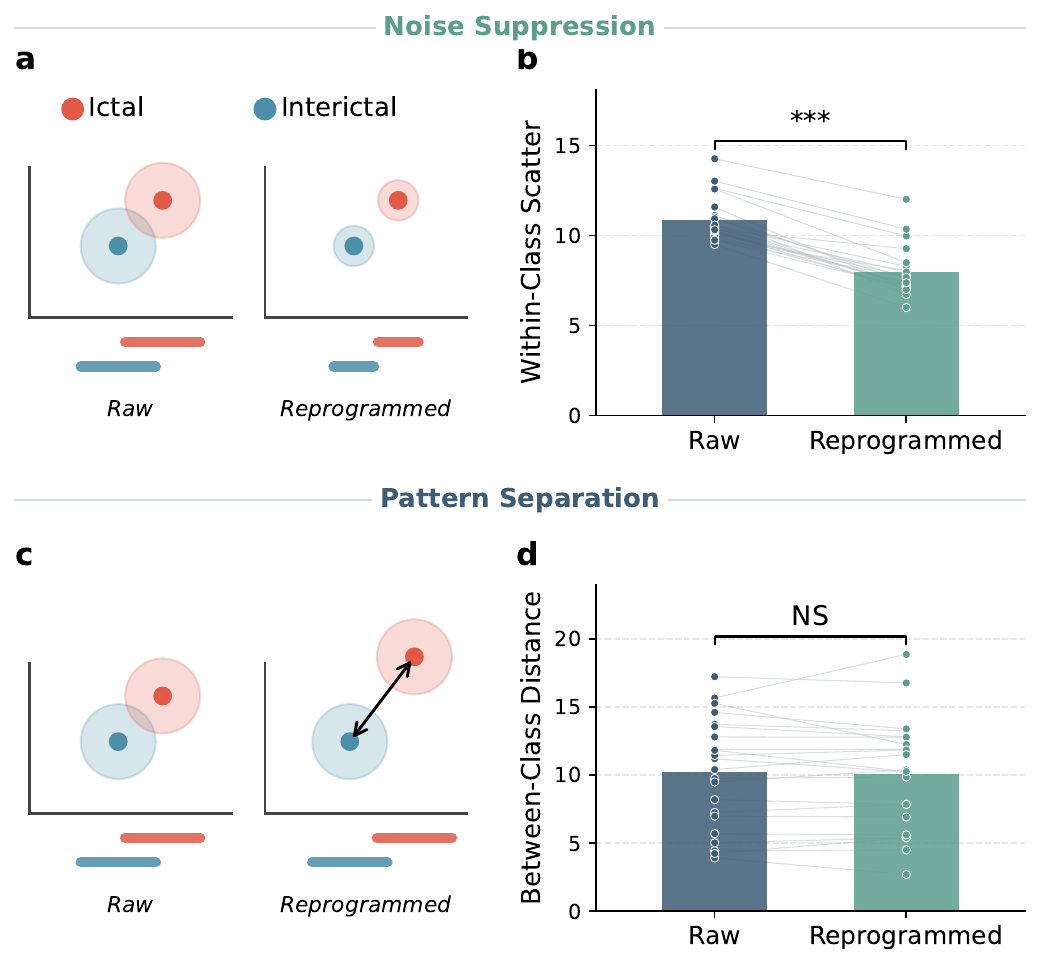}
\caption{Reprogramming behaves as a denoiser rather than a pattern amplifier. Top: within-class scatter falls sharply under reprogramming. Bottom: the between-class distance separating the ictal and interictal centroids is unchanged. The separability gain therefore comes entirely from suppressing within-class variability, not from pushing the classes apart.}
\label{fig:denoiser}
\end{figure}

\subsection{Main Results}
\label{subsec:main}

\paragraph{The iEEG prior improves every host model on every dataset.} In Figure~\ref{fig:main}, reprogramming through the iEEG-trained prior improves the host model in every one of the sixteen model-dataset combinations, and it does so on all six metrics rather than on a favorable subset. The average improvement is roughly $0.08$ on binary F1, sensitivity, precision, and the Matthews correlation coefficient, and a smaller $0.04$ on balanced accuracy and AUROC. Because AUROC is independent of the decision threshold and still rises in every cell, the prior improves how the model ranks ictal against interictal windows and not merely where it places its threshold. The improvement is statistically significant, not a chance fluctuation. Where a per-patient score distribution exists, a cross-patient Wilcoxon signed-rank test confirms it, with the more strongly affected metrics reaching higher significance, indicating a systematic pattern. The gain is also structured. It is larger for the three time-domain hosts than for the spectral-input DCRNN, and it grows as the raw signal becomes harder to classify. On Siena, the hardest dataset, the time-domain hosts roughly double their F1 while DCRNN changes little. This inverse coupling between raw difficulty and gain recurs across datasets, indicating that the prior contributes most when the host can extract the least from the raw signal.

\paragraph{The gain comes from the iEEG, not the architecture itself.} To separate the contribution of the iEEG training signal from that of the architecture alone, we additionally reprogram each input through an identical prior fitted on scalp EEG rather than iEEG, which matches the iEEG prior in architecture and capacity and differs only in the modality of its fitting corpus. This scalp-trained control does not reproduce the effect. It performs at the raw level throughout the grid and frequently a little below it, which is exactly what a faithful reconstruction produces. Fitted on scalp EEG and then applied back to scalp EEG, the control merely re-encodes its own training modality and returns the input essentially unchanged, so the host still sees the raw signal, with the small shortfall reflecting only the lossy bottleneck. Passing the signal through this architecture is therefore not what helps. Re-expressing it specifically through a codebook learned from cleaner intracranial recordings is. Because the two priors are matched in capacity and preprocessing and the host models, protocols, and evaluation are held fixed, this comparison attributes the downstream gain to the iEEG training corpus itself, which is the unpaired cross-modal prior the method is built around.

\subsection{A Representation-Geometry View of Reprogramming}
\label{subsec:mechanism}

We ask a mechanistic question: what does the frozen prior do to the scalp signal, and why does it
help? Because no clean-signal ground truth exists for scalp EEG (Section~\ref{sec:setup}), we
cannot directly measure how much noise the prior removes. We therefore adopt a label-only,
classifier-free probe grounded in signal-detection theory~\cite{green1966signal} that quantifies how
separable seizure (ictal) and background (interictal) activity is before and after reprogramming,
decomposing any change into its geometric causes.
We use six classic clinical morphology descriptors, namely line length~\cite{esteller2001line}, the
three Hjorth parameters~\cite{hjorth1970eeg}, kurtosis~\cite{alam2013detection}, and zero-crossing
rate~\cite{pyrzowski2021zero}, of which line length is the canonical seizure biomarker~\cite{esteller2001line}. In this
feature space,
with $\mu_{\mathrm{ict}}$ and $\mu_{\mathrm{int}}$ the ictal and interictal class centroids and
$\sigma^2_{\mathrm{ict}}, \sigma^2_{\mathrm{int}}$ their within-class spreads, we measure
separability by a standardized mean difference (Cohen's $d$)~\cite{cohen1988statistical}, equivalently
the signal-detection sensitivity index~\cite{green1966signal}, $d = \lVert \mu_{\mathrm{ict}} -
\mu_{\mathrm{int}} \rVert / \sqrt{\tfrac{1}{2}(\sigma^2_{\mathrm{ict}} + \sigma^2_{\mathrm{int}})}$,
the ratio of the between-class centroid distance to the within-class spread.
Under reprogramming, $d$ in the joint six-feature space rises from $0.93$ to $1.26$ (Wilcoxon
signed-rank $p<10^{-6}$), and the gain is broad rather than driven by any single descriptor, with
all six morphology features improving significantly on their own (Figure~\ref{fig:perfeat}).
This index can rise in two ways, by pushing the class centroids apart (Figure~\ref{fig:denoiser}c)
or by tightening each class (Figure~\ref{fig:denoiser}a), and measuring the two terms separately
shows that the between-class distance is essentially unchanged (Figure~\ref{fig:denoiser}d), so the
entire $d$ gain is attributable to a reduction of within-class scatter (Figure~\ref{fig:denoiser}b). Reprogramming does not push seizure and background apart. It
makes each class more internally consistent by suppressing within-class variability, the signature
of a denoiser rather than a signal amplifier. Nuisance variation that inflates within-class scatter
without carrying class-discriminative information is attenuated, while the cortical structure that
separates the classes is preserved.

\begin{figure}[t]
\centering
\includegraphics[width=\linewidth]{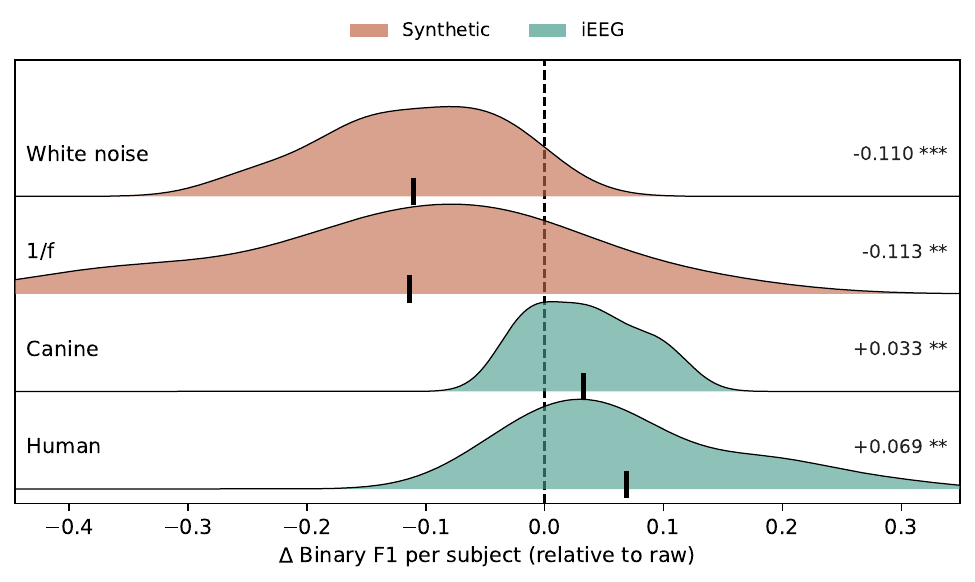}
\caption{Effect of the corpus on which the prior is fitted. Each row is the density curve of the per-subject changes relative to raw. Vertical ticks mark the mean and stars mark a per-subject Wilcoxon signed-rank test against raw.}
\label{fig:ridgeline}
\end{figure}

\subsection{Isolating the Source of the Gain}
\label{subsec:source}

The scalp-trained control of Section~\ref{subsec:main} established that the gain comes from the iEEG rather than from the architecture. We take this one step further and ask which property of that corpus carries it. The prior might transfer (i) genuine neural structure that intracranial recording captures with high fidelity, or (ii) only the low-level spectral statistics of iEEG, in particular the approximately $1/f$ falloff its power spectrum shares with other field potentials~\cite{pritchard1992fractal, miller2009powerlaw, donoghue2020parameterizing, he2014scalefree}, in which case any signal with a similar spectrum would serve equally well and the prior would reduce to a generic spectral filter with no specifically neural content.

We separate these two accounts by refitting the prior on two families of corpora that dissociate spectral statistics from neural content. The first contains synthetic signals with no neural structure, namely white noise with a flat spectrum and a $1/f$ signal whose spectrum is matched to the power falloff of brain field potentials. The second contains real iEEG from a non-human species, namely canine~\cite{brinkmann2016canine}, together with the human iEEG used throughout. This non-human corpus is detailed in Appendix~\ref{app:corpora}. If the transferable structure is genuinely neural then recordings from another species should carry some of it.

Figure~\ref{fig:ridgeline} shows a clean split between the two families. Both synthetic corpora fall significantly below raw, which rules out account (ii). The real recordings help instead, and the effect is not limited to human iEEG. This is because mammalian brain rhythms are largely conserved across species~\cite{buzsaki2013scaling}, though the canine corpus helps less than human iEEG.

\section{Conclusion}
In this paper, we propose \method, a cross-modal reprogramming framework that transfers a prior learned from intracranial EEG to scalp EEG without any paired data. \method fits a channel-agnostic codebook model on iEEG alone and freezes it, so that any scalp signal passed through it is re-expressed as a sequence of cortical-signal atoms and consumed by any downstream model without fine-tuning. Experiments across recordings from four hospitals and host models spanning the major architecture families show that \method consistently and significantly improves downstream seizure detection. Extending the prior to non-pathological tasks will be explored in future work.

\section{Limitations and Ethical Considerations}
\label{sec:ethics}

Because the clinical iEEG comes from epilepsy patients, the prior
helps most on tasks that draw on the same structure, such as seizure detection, and
its gain on brain-computer interface tasks does not yet reach statistical significance. All records used in this work come from existing public corpora that
were de-identified and released by their source institutions under those institutions'
ethical approval and data use agreements. The method is agnostic to subject
demographics and therefore introduces no demographic bias of its own, and it decodes
no private mental content, which keeps its misuse surface limited.

\section{Generative AI Usage}
\label{sec:genai}

Generative AI was used only to improve the clarity and language of the writing and
to assist with LaTeX formatting.

\begin{acks}
This work was supported by the NVIDIA Academic Grant in Higher Education and Developer Program.
\end{acks}

\newpage

\bibliographystyle{ACM-Reference-Format}
\bibliography{references}
\balance

\newpage

\appendix
\section*{APPENDIX}

\section{Method Details}
\label{app:loss}

We detail the composite generator loss summarized in Section~\ref{subsec:objective}, where the total generator loss decomposes as $\mathcal{L}_{\mathrm{gen}} = \mathcal{L}_{\mathrm{recon}} + \mathcal{L}_{\mathrm{adv}} + \mathcal{L}_{\mathrm{commit}}$.

\paragraph{Reconstruction component.} The reconstruction component aggregates three atomic losses. A time-domain $L_1$ term $\ell_t = \|x - \hat{x}\|_1$ enforces pointwise waveform fidelity. A multi-resolution spectral term
\begin{equation}
\ell_s \;=\; \sum_{i \in \mathcal{I}} \tfrac{1}{|S_i(x)|}\!\left(\|S_i(x) - S_i(\hat{x})\|_1 + \gamma\,\|S_i(x) - S_i(\hat{x})\|_2\right)
\end{equation}
enforces frequency-domain fidelity across temporal scales, where $S_i$ is the STFT magnitude at window size $2^i$, $\mathcal{I} = \{5, 6, \dots, 11\}$, and $\gamma$ balances the two norms. A relative line-length term $\ell_l$ captures the local first-difference variation of the waveform over sliding windows, preserving fine-grained temporal structure. The aggregated reconstruction component is $\mathcal{L}_{\mathrm{recon}} = \lambda_t \ell_t + \lambda_s \ell_s + \lambda_l \ell_l$.

\paragraph{Adversarial component.} A multi-scale STFT discriminator $\mathcal{D}$ provides two atomic losses. The hinge generator loss is $\ell_g = \mathbb{E}_x\!\left[\max(0,\, 1 - \mathcal{D}(\hat{x}))\right]$. The feature-matching loss $\ell_f$ is the $L_1$ distance between the intermediate feature maps of $\mathcal{D}$ on $x$ and $\hat{x}$, summed over layers and normalized in each layer by the $L_1$ norm of the real-signal feature map. The aggregated adversarial component is $\mathcal{L}_{\mathrm{adv}} = \lambda_g \ell_g + \lambda_f \ell_f$. The discriminator $\mathcal{D}$ is not part of $\mathcal{L}_{\mathrm{gen}}$ and is instead trained by its own objective in alternation with $f_\theta$,
\begin{equation}
\mathcal{L}_{\mathrm{disc}} \;=\; \mathbb{E}_x\!\left[\max(0,\, 1 - \mathcal{D}(x))\right] + \mathbb{E}_x\!\left[\max(0,\, 1 + \mathcal{D}(\hat{x}))\right].
\end{equation}

\paragraph{Quantizer commitment.} The commitment term anchors encoder outputs to current codebook entries via
\begin{equation}
\mathcal{L}_{\mathrm{commit}} \;=\; \lambda_q \sum_{m=1}^{M} \|z^{(m)} - \mathrm{sg}[q^{(m)}]\|_2^2,
\end{equation}
where $\mathrm{sg}[\cdot]$ is the stop-gradient operator. The stop-gradient on $q^{(m)}$ blocks gradient flow into the codebook, which is instead trained through the EMA updates described in Section~\ref{subsec:rvq}. The commit loss therefore only updates the encoder, pulling its outputs toward the current codebook entries.

\paragraph{Gradient-norm loss balancing.} Rather than tuning the relative weights of the reconstruction and adversarial terms by hand, we set them with the gradient-norm balancer of EnCodec~\cite{defossez2022encodec}. It rescales each term by a running estimate of its own gradient norm, so that each contributes a fixed fraction of the training gradient rather than an amount set by its raw magnitude, keeping any single loss from dominating despite the disparate scales of the time-domain, spectral, and adversarial terms. The commitment loss is applied directly and is not balanced.

\section{Experiment Details}
\label{app:exp}

\subsection{Model Configuration}
\method is the channel-independent encoder-decoder of Section~\ref{sec:method} with a residual vector-quantized bottleneck. The encoder applies six stride-2 downsampling stages with latent dimension $d{=}64$, and the bottleneck uses $M{=}4$ residual codebooks of $K{=}256$ entries each, a budget of 32 bits per latent step. It is fitted on iEEG for 96k steps with the gradient-norm-balanced objective of Section~\ref{subsec:objective}, after which all parameters are frozen. The backbone $f_\theta$ and the discriminator $\mathcal{D}$ are trained with separate Adam optimizers, each under a cosine learning-rate schedule with linear warmup.

\subsection{Multi-Scale STFT Discriminator}
The discriminator $\mathcal{D}$ of Section~\ref{subsec:discriminator} uses $J{=}5$ sub-discriminators at $(n_{\mathrm{fft}}, h, w_{\mathrm{win}})$ resolutions (2048, 512, 2048), (1024, 256, 1024), (512, 128, 512), (256, 64, 256), and (128, 32, 128), powers of 2 spanning the time-frequency range relevant for cortical signals at our sampling rate. It is a training-time auxiliary and is discarded once fitting completes.

\subsection{Preprocessing}
Preprocessing differs between the intracranial corpus on which the prior is fitted and the scalp corpora on which it is evaluated, because the two modalities do not admit the same channel conventions.

\paragraph{Intracranial corpus for prior fitting.} SWEC is recorded at 512\,Hz and is kept at that rate, which is the rate at which \method operates. Intracranial montages are patient-specific in electrode count and anatomical placement and follow no standard electrode-position convention, so no standard re-referencing scheme is defined for them and we retain every channel in its native recording reference. Because electrode counts differ across patients, each window draws a random subset of 36 channels, the smallest electrode count among the 16 patients, which fixes the batch shape while leaving every channel able to be selected. Recordings are segmented into 4\,s windows at 1\,s stride. Fitting is self-supervised on the waveform alone, so no labels or class balancing are applied.

The non-human corpus of Appendix~\ref{app:corpora} follows the same protocol after polyphase resampling to 512\,Hz from its native rate, drawing 15 channels per window for the canine corpus, whose recordings carry 16 electrodes in total.

\paragraph{Scalp corpora for downstream evaluation.} Scalp recordings do follow the standard 10--20 electrode-position convention, so we reduce every dataset to the same 18-channel longitudinal bipolar montage, known clinically as the double banana, which places the four datasets in a common channel space despite their differing native electrode subsets and reference schemes. TUSZ, Siena, and Beirut are distributed with monopolar references and are re-referenced to this montage, then anti-alias filtered with a zero-phase fourth-order Butterworth lowpass at 64\,Hz and resampled to 256\,Hz from their native rates. CHB-MIT is distributed already bipolar-derived at 256\,Hz and requires neither step. Every recording is then segmented into 4\,s windows at 1\,s stride. A window is labeled ictal if it overlaps a clinician-annotated seizure by at least 50\%, and interictal otherwise. During training we downsample the background class to a 5:1 background-to-seizure ratio, while the test distribution keeps its native imbalance. Because the prior operates at 512\,Hz, a window is resampled to 512\,Hz for reprogramming and returned to 256\,Hz before it reaches the host model, so the host receives the same sampling rate and channel layout in the raw and the reprogrammed condition. The scalp-fitted control of Section~\ref{subsec:main} follows this identical resampling path, so the round-trip is shared across the reprogrammed conditions rather than specific to the iEEG prior.

\subsection{Non-Human iEEG Corpora}
\label{app:corpora}
The canine corpus is long-term intracranial EEG from five dogs with naturally occurring epilepsy, recorded through the chronically implanted NeuroVista seizure advisory system, an ambulatory device that transmits sixteen channels of iEEG by wireless telemetry at 400\,Hz. Each dog carries sixteen subdural electrodes in a bilaterally symmetric arrangement of paired four-contact strips running from anterior to posterior over each hemisphere. These five are the animals among eight implanted whose recordings reached sufficient quality and seizure count for analysis, and their continuous monitoring spans months, from $1{,}920$ to $8{,}208$ hours across the five dogs. The recordings were released for the 2014 American Epilepsy Society Seizure Prediction Challenge, hosted on Kaggle with sponsorship from the American Epilepsy Society, the Epilepsy Foundation, and the National Institutes of Health~\cite{brinkmann2016canine}.

\section{Additional Results}
\label{app:additional}

\begin{table}[htbp]
\centering
\caption{Downstream seizure detection on CHB-MIT with EEGWaveNet as the host, comparing \method against EEG denoising front-ends.}
\label{tab:denoising}
\setlength{\tabcolsep}{1.5pt}
\begin{tabular}{lcccccc}
\toprule
Front-end & BF1 & Sn & Pr & BAcc & MCC & AUROC \\
\midrule
Raw                  & 0.681 & 0.717 & 0.706 & 0.852 & 0.688 & 0.955 \\
Wavelet thresholding & 0.657 & 0.733 & 0.654 & 0.852 & 0.662 & 0.955 \\
ASR                  & 0.667 & 0.732 & 0.690 & 0.848 & 0.674 & 0.953 \\
ICA                  & 0.661 & 0.697 & 0.705 & 0.841 & 0.671 & 0.956 \\
EEGdenoiseNet (EOG)  & 0.604 & 0.651 & 0.630 & 0.817 & 0.611 & 0.945 \\
EEGdenoiseNet (EMG)  & 0.626 & 0.661 & 0.650 & 0.823 & 0.631 & 0.939 \\
\midrule
\method              & \textbf{0.749} & \textbf{0.776} & \textbf{0.774} & \textbf{0.882} & \textbf{0.754} & \textbf{0.972} \\
\bottomrule
\end{tabular}
\end{table}

\subsection{Comparison with Classical Denoising Front-Ends}
\label{app:denoising}

We compare \method against four EEG denoising front-ends that span the families into which EEG denoising falls. Wavelet thresholding~\cite{donoho1994ideal} decomposes each channel over a wavelet basis, estimates the noise level from the finest scale, and shrinks the coefficients that fall below a universal threshold, so its dictionary is fixed and analytic rather than learned from data. Artifact subspace reconstruction (ASR)~\cite{mullen2015real} and independent component analysis (ICA)~\cite{hyvarinen2000independent} instead exploit structure across channels. The former calibrates on the low-variance portion of each recording and reconstructs the components whose variance exceeds a threshold, while the latter decomposes each recording into maximally independent components and discards those whose kurtosis identifies them as artifactual~\cite{delorme2007enhanced}.

The three methods above are unsupervised. We further include EEGdenoiseNet~\cite{zhang2021eegdenoisenet}, a supervised deep denoiser, whose Novel CNN we train on the benchmark's paired data at controlled signal-to-noise ratios. Following the benchmark's own protocol, which fits a separate network for each artifact type, we train one model for ocular and one for myogenic contamination and deploy each as a front-end.

Each denoiser replaces \method at the same position and returns a signal of the same shape, so the host, its training protocol, the cross-validation split, and the metrics all follow Section~\ref{subsec:setup} unchanged and the front-end is the only thing that varies. Table~\ref{tab:denoising} reports the outcome. Not one of the five improves on the raw signal, and each falls below it on both binary F1 and the Matthews correlation coefficient. The three unsupervised methods hold AUROC within $0.002$ of the raw value, so they barely alter how ictal and interictal windows are ranked, while the supervised denoiser lowers it. Supervised denoising is characteristically fitted and scored against a proxy target that cannot itself be guaranteed clean, with relative root mean squared error and the correlation coefficient measuring how closely an output matches that proxy. The performance such work reports is throughout this agreement with the proxy, and the downstream task the denoising is ultimately meant to serve goes unexamined. \method is the only front-end that improves on raw, and it does so on every metric. Cleaning the scalp signal by generic means is therefore not sufficient on its own. What the host benefits from is the re-expression of the signal through a codebook learned from intracranial recordings.

\begin{figure*}[htbp]
\centering
\includegraphics[width=\linewidth]{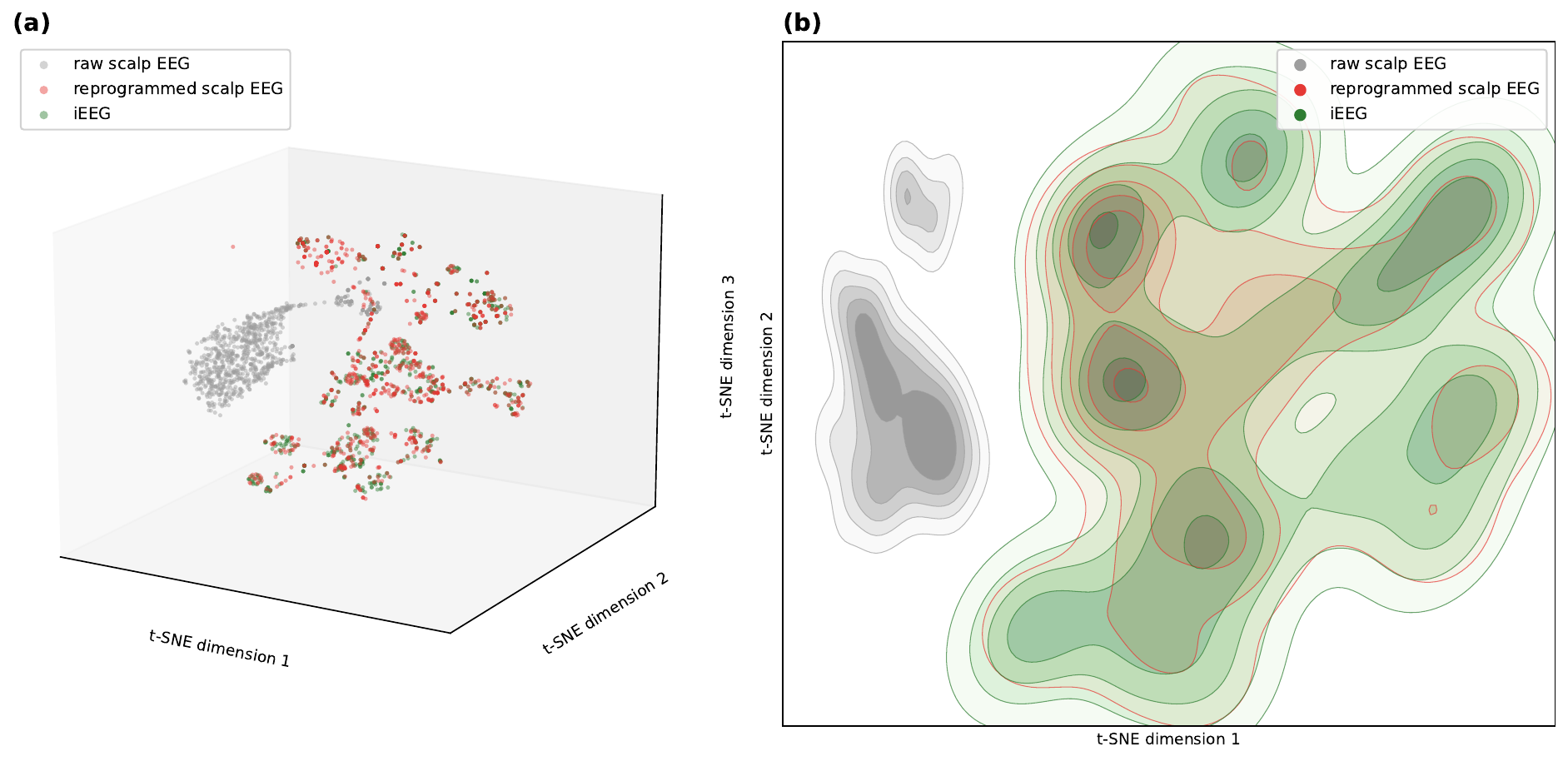}
\caption{Representation-manifold view of reprogramming. (a) Individual windows in three dimensions and (b) class-density contours in two dimensions, for raw scalp EEG, reprogrammed scalp EEG, and iEEG, in the \method\ latent representation embedded with t-SNE.}
\label{fig:manifold}
\end{figure*}

\subsection{Replication across Additional Clinical Centers}
\label{app:mccorpus}

\begin{table}[htbp]
\centering
\caption{Downstream seizure detection on CHB-MIT with EEGWaveNet as the host, for a prior fitted on the Multi-Center (MC) iEEG corpus.}
\label{tab:mccorpus}
\setlength{\tabcolsep}{3pt}
\begin{tabular}{lcccccc}
\toprule
Fitting corpus & BF1 & Sn & Pr & BAcc & MCC & AUROC \\
\midrule
Raw          & 0.681 & 0.717 & 0.706 & 0.852 & 0.688 & 0.955 \\
\method (MC) & \textbf{0.721} & \textbf{0.770} & \textbf{0.742} & \textbf{0.874} & \textbf{0.728} & \textbf{0.969} \\
\bottomrule
\end{tabular}
\end{table}

The scalp-trained control of Section~\ref{subsec:main} attributes the gain to the iEEG training corpus, and Section~\ref{subsec:source} traces it to genuine neural structure rather than spectral statistics. A further question is whether the effect is a property of the particular corpus on which the prior is fitted or of high-fidelity human iEEG in general. We address it by refitting the prior on human iEEG from additional clinical centers and deploying it under the identical protocol.

The Multi-Center (MC) iEEG dataset~\cite{li2021fragility} aggregates intracranial recordings around ictal events from four clinical centers, all distinct from the SWEC source.
We follow the SWEC preprocessing of Appendix~\ref{app:exp}. The backbone, bottleneck, training schedule, and channel-wise deployment are all held identical to the SWEC prior, so the fitting corpus is the only thing that varies.

Table~\ref{tab:mccorpus} reports the outcome on CHB-MIT with EEGWaveNet as the host. The MC-fitted prior improves on the raw signal across all six metrics, raising binary F1 from $0.681$ to $0.721$. A cross-patient Wilcoxon signed-rank test on per-patient F1 finds it statistically indistinguishable from the SWEC prior ($p=0.59$).

\subsection{Representation-Manifold View of Reprogramming}
\label{app:manifold}

A natural question is what reprogramming does to a scalp signal in the representation that \method assigns to it. We examine this through the codebook representation. We collect signal windows from three sources, namely raw scalp EEG, the same scalp EEG after reprogramming, and iEEG, encode each window with \method, and embed the resulting latent vectors with t-SNE.

Figure~\ref{fig:manifold} shows a consistent picture across the per-window scatter and the class-density view. Raw scalp EEG forms a region of its own that is well separated from iEEG, whereas reprogrammed scalp EEG intermingles with iEEG and shares the same region of the representation. Because the codebook is fitted on iEEG alone and never sees scalp EEG, this is the behavior expected if reprogramming re-expresses a scalp signal through the iEEG-learned codebook rather than leaving it in its native scalp representation.

\section{Deployment Cost of the Frozen Front-End}
\label{app:cost}

\method is inserted upstream of a host model and left frozen. We report the resources it requires at inference, from a raw scalp window entering it to the reprogrammed window leaving it.

\paragraph{Size.} Table~\ref{tab:cost-size} reports what is deployed. The multi-scale STFT discriminator of Section~\ref{subsec:discriminator} is an auxiliary that is discarded once the prior is fitted, so the deployed model is an encoder, a decoder, and the residual codebook, together $2.24$M parameters.

\begin{table}[htb]
\centering
\caption{Size of the deployed \method front-end. The discriminator is used only while the prior is fitted and is not part of what is deployed.}
\label{tab:cost-size}
\setlength{\tabcolsep}{4pt}
\begin{tabular}{lr}
\toprule
Component & Parameters \\
\midrule
Encoder                                 & 1{,}055{,}432 \,($1.06$M) \\
Decoder                                 & 1{,}055{,}369 \,($1.06$M) \\
Codebook ($M{=}4$, $K{=}256$, $d{=}64$) & 132{,}100 \,($0.13$M) \\
\midrule
\textbf{Deployed front-end}             & \textbf{2{,}242{,}901} \,(\textbf{$2.24$M}) \\
\midrule
\textit{Discriminator (discarded)}      & \textit{1{,}004{,}298} \,(\textit{$1.00$M}) \\
\bottomrule
\end{tabular}
\end{table}

\paragraph{Compute.} Table~\ref{tab:cost-compute} reports the cost of reprogramming a full $18$-channel window, which is the unit a host consumes. The path we time is the deployed one in full.

\begin{table}[htb]
\centering
\caption{Compute cost of reprogramming. Latency is the median over six runs on a single CPU thread of an Intel Xeon w3-2423, with the observed range in parentheses. The real-time factor is compute time divided by signal duration, so a value below one is faster than real time.}
\label{tab:cost-compute}
\setlength{\tabcolsep}{4pt}
\begin{tabular}{lr}
\toprule
Quantity & Value \\
\midrule
\multicolumn{2}{l}{\textit{Per $18$-channel $4$\,s window, one CPU thread}} \\
Latency                                 & $175$\,ms \,($148$--$234$) \\
Real-time factor                        & $0.044$ \,($0.037$--$0.059$) \\
Time to reprogram one hour of recording & $157$\,s \\
\bottomrule
\end{tabular}
\end{table}

A single CPU thread reprograms a $4$\,s window of an $18$-channel montage in about $0.18$\,s, more than twenty times faster than real time, so the front-end can sit in a live recording path without becoming the bottleneck. Reprogramming a full hour of recording offline takes under three minutes on that same thread, against the hour the recording itself took to acquire. Our measurements were taken on a machine shared with unrelated concurrent jobs, which is why we report a range, but the conclusion holds anywhere in that range, since even the slowest observation leaves the front-end seventeen times faster than real time.

The deployed front-end holds $2.24$M parameters and reprograms a window more than twenty times faster than real time on a single thread. A single-board computer such as the Raspberry Pi 5, which carries four Arm Cortex-A76 cores at $2.4$\,GHz and up to $16$\,GB of memory~\cite{raspberrypi5brief}, therefore accommodates it. Hardware of this class is already widely used for clinical and ambulatory health monitoring~\cite{baraneedharan2025raspberrypi}. Reprogramming is therefore practical to deploy at the clinical edge.

\end{document}